\documentclass[conference,a4paper]{IEEEtran}
\IEEEoverridecommandlockouts

\usepackage[hidelinks]{hyperref}
\usepackage[cmex10]{amsmath}%American Math Society(AMS) math formatting
\usepackage{amssymb,amsfonts}%AMS extra symbols and fonts
\usepackage{dblfloatfix}%fix double column figure ordering and placement

\usepackage[ruled,vlined]{algorithm2e}
\usepackage{graphicx}
\graphicspath{{Figures/PDF/}{Figures/PNG/}}
\usepackage{multirow}

\usepackage{booktabs}
\usepackage{siunitx}
\usepackage[numbers,compress]{natbib}
\usepackage{texnames}
\usepackage{bm,bbm}
\usepackage{orcidlink}
\usepackage{comment}

\begin{document}

% \title{\uppercase{Analysis and Classification of Weed Management Practices in Agricultural Fields Using Remote Sensing and Machine Learning}
% \title{\uppercase{Analysis and Classification of Weed Management Practices in Permanent Crop Fields Using Remote Sensing and Machine Learning}
% \title{\uppercase{Mapping of Weed Management Practices in Permanent Crops Using Remote Sensing}
\title{\uppercase{Mapping of Weed Management Methods in Orchards using 
Sentinel-2 and PlanetScope Data}
\thanks{\textcopyright\ 2025 IEEE. Published in the 2025 IEEE International Geoscience and Remote Sensing Symposium (IGARSS 2025), scheduled for 3 - 8 August 2025 in Brisbane, Australia. Personal use of this material is permitted. However, permission to reprint/republish this material for advertising or promotional purposes or for creating new collective works for resale or redistribution to servers or lists, or to reuse any copyrighted component of this work in other works, must be obtained from the IEEE. Contact: Manager, Copyrights and Permissions / IEEE Service Center / 445 Hoes Lane / P.O. Box 1331 / Piscataway, NJ 08855-1331, USA. Telephone: + Intl. 908-562-3966.}
\thanks{This version is the accepted manuscript submitted to arXiv. The final version will be published in the Proceedings of IGARSS 2025 and available via IEEE Xplore. For citation, please refer to the published version in IGARSS 2025.}
}
% in order to target for MultiSource theme
% Earth Observation Data}}

% \author{	\IEEEauthorblockN{Ioannis Kontogiorgakis \orcidlink{0000-0002-8002-5341}}
% \IEEEauthorblockA{\textit{National Observatory of Athens}\\
% 		6140 Athens, Greece\\
% 		ikontog@noa.gr}

% \and

% \IEEEauthorblockN{Iason Tsardanidis\orcidlink{0000-0002-5283-7350}}
% \IEEEauthorblockA{\textit{National Observatory of Athens}\\
% 		010021 Athens, Greece\\
% 		j.tsardanidis@noa.gr}
	
% \and

% \IEEEauthorblockN{Ilias Tsoumas\orcidlink{0000-0002-9185-1382}}
% \IEEEauthorblockA{\textit{National Observatory of Athens}\\
% 		1688 Athens, Greece\\
% 		i.tsoumas@noa.gr}
% }

\author{
    Ioannis Kontogiorgakis\textsuperscript{1},
    Iason Tsardanidis*\textsuperscript{1},
    \thanks{*Correspondence to  \href{mailto:j.tsardanidis@noa.gr}{j.tsardanidis@noa.gr}}
    Dimitrios Bormpoudakis\textsuperscript{1},
    Ilias Tsoumas\textsuperscript{1,2},\\
    Dimitra A. Loka\textsuperscript{3},
    Christos Noulas\textsuperscript{3}, 
    Alexandros Tsitouras\textsuperscript{3} and
    Charalampos Kontoes\textsuperscript{1}\\[1em]
    
    \textsuperscript{1}BEYOND EO Centre, IAASARS, National Observatory of Athens, Athens, Greece\\
    \textsuperscript{2}Artificial Intelligence, Wageningen University \& Research, The Netherlands\\
    \textsuperscript{3}Institute of Industrial and Forage Crops, Hellenic Agricultural Organization (ELGO) “DIMITRA”, Larissa, Greece
}

\maketitle

\begin{abstract}
    Effective weed management is crucial for improving agricultural productivity, as weeds compete with crops for vital resources like nutrients and water. 
    Accurate maps of weed management methods are essential for policymakers to assess farmer practices, evaluate impacts on vegetation health, biodiversity, and climate, as well as ensure compliance with policies and subsidies.
    However, monitoring weed management methods is challenging as they commonly rely on ground-based field surveys, which are often costly, time-consuming and subject to delays.
    In order to tackle this problem, we leverage earth observation data and Machine Learning (ML). Specifically, we developed separate ML models using Sentinel-2 and PlanetScope satellite time series data, respectively, to classify four distinct weed management methods (Mowing, Tillage, Chemical-spraying, and No practice) in orchards. The findings demonstrate the potential of ML-driven remote sensing to enhance the efficiency and accuracy of weed management mapping in orchards.
    % Specifically, we developed an ML approach for mapping four distinct weed management methods (\textit{Mowing}, \textit{Tillage}, \textit{Chemical-spraying}, and \textit{No practice}) in orchards using satellite image time series (SITS) data from two different sources: Sentinel-2 (S2) and PlanetScope (PS). The findings demonstrate the potential of ML-driven remote sensing to enhance the efficiency and accuracy of weed management mapping in orchards.

\end{abstract}

% \begin{IEEEkeywords}
% 	Weed Management, Remote Sensing, PlanetScope, Sentinel-2, Machine Learning.
% \end{IEEEkeywords}

\section{Introduction}

Weed management is considered essential in agriculture, directly influencing crop productivity, resource efficiency and environmental sustainability \cite{monteiro2022sustainable}, \cite{bajwa2015nonconventional}. In orchards, effective weed management is particularly critical, as weeds compete with trees for water and nutrients, reduce yields, and increase vulnerability to pests and diseases\cite{parker2003water}.

The three primary weed management methods in orchards are mowing, tillage, and chemical spraying via herbicides, each offering distinct advantages and challenges. Mowing is typically performed using mechanical mowers or trimmers to cut weeds to a manageable height, maintaining ground cover. Tillage involves the use of plows, harrows, or rotary tillers to physically disrupt the soil, uprooting weeds and incorporating organic matter. Chemical spraying with herbicides is implemented through the use of specialized equipment such as boom sprayers or handheld applicators, allowing precise application of selective or non-selective weed killers to manage weed species.

Traditionally, weed management practices — such as mowing, mechanical tilling, and chemical spraying — are monitored through field surveys or manual records. Furthermore, Unmanned Aerial Vehicles (UAVs) equipped with multispectral sensors have emerged as a promising tool for weed management tasks, enabling the efficient collection of high-resolution spatial and spectral data \cite{Imanni2023Multispectral, Esposito2021Drone}. However, both field surveys and UAV-based methods may incur high costs and/or legal/safety concerns (over UAV flights) present a significant barrier to widespread adoption and scalability. 

Earth Observation (EO) data have revolutionized agriculture monitoring, providing temporal and spatial insights in vegetation \cite{sishodia2020applications}. Satellite imagery enables the capture of precise variations in vegetation health and coverage. The Sentinel-2 (S2) constellation is widely used for ground-soil monitoring due to its high temporal resolution and multi-spectral capabilities. 
However, its limited spatial resolution can pose challenges for various applications, such as detailed weed management mapping, particularly in orchards where the overlapping tree crowns and ground pixels complicate accurate analysis and classification.
PlanetScope (PS) imagery \cite{planet2024}, with its higher spatial and temporal resolution, has been demonstrated in various applications to overcome the limitations of mid-resolution satellites like S2. Its finer spatial detail enables more accurate discrimination within fields, reducing the prevalence of mixed pixel (mixel) issues \cite{mudereri2019comparative}. Studies have shown that EO imagery, combined with advanced machine learning (ML) techniques, can facilitate precise mapping and classification of agricultural practices \cite{tsardanidis_grasslands,gao2018fusion,su2022spectral,fragassa2023new}.

In this study, we propose a novel ML-based approach to monitor and classify four distinct weed management methods — \textit{Mowing}, \textit{Tillage}, \textit{Chemical spraying} and \textit{No practice} - utilizing multi-source EO data. By leveraging the spectral and temporal capabilities of S2 and PS imagery, our methodology aims to provide an automated, scalable solution for assessing and differentiating these methods in orchards. This approach not only reduces reliance on traditional labor-intensive on-ground monitoring but also provides a useful mapping tool for agro-ecosystem researchers, Common Agricultural Policy (CAP) monitoring agencies and relevant policymakers.

\begin{comment}
    
This template is aligned with the good practices stated by \citet{EditorialGRSL2015}, which are a step towards a reproducible article \citep{ABadgingSystemforReproducibilityandReplicabilityinRemoteSensingResearch}.
We used the official IEEE Conference Proceedings template available here \url{https://www.ieee.org/conferences/publishing/templates.html} to produce this simplified version for IGARSS~2025.
This template uses \BibTeX, a convenient way of managing references that simplifies the formatting and automatically selects the bibliographic entries cited in the text.

A basic knowledge of \LaTeX\ and \BibTeX\ is required.

\end{comment}

% \begin{figure}[]
%     \centering
%     % \includegraphics[width=0.45\textwidth]
%     \includegraphics[width=\columnwidth]
%     {Figures/PNG/fig1.png}
%     \caption{Our area of interest in Thessaly, Greece. Yellow points represent the fields where we identified a weed management method.}
%     \label{fig:2.1}
% \end{figure}

\section{Data Collection}

\subsection{Point Dataset}

\begin{figure}[ht!]
    \centering
    \includegraphics[width=\columnwidth]
    {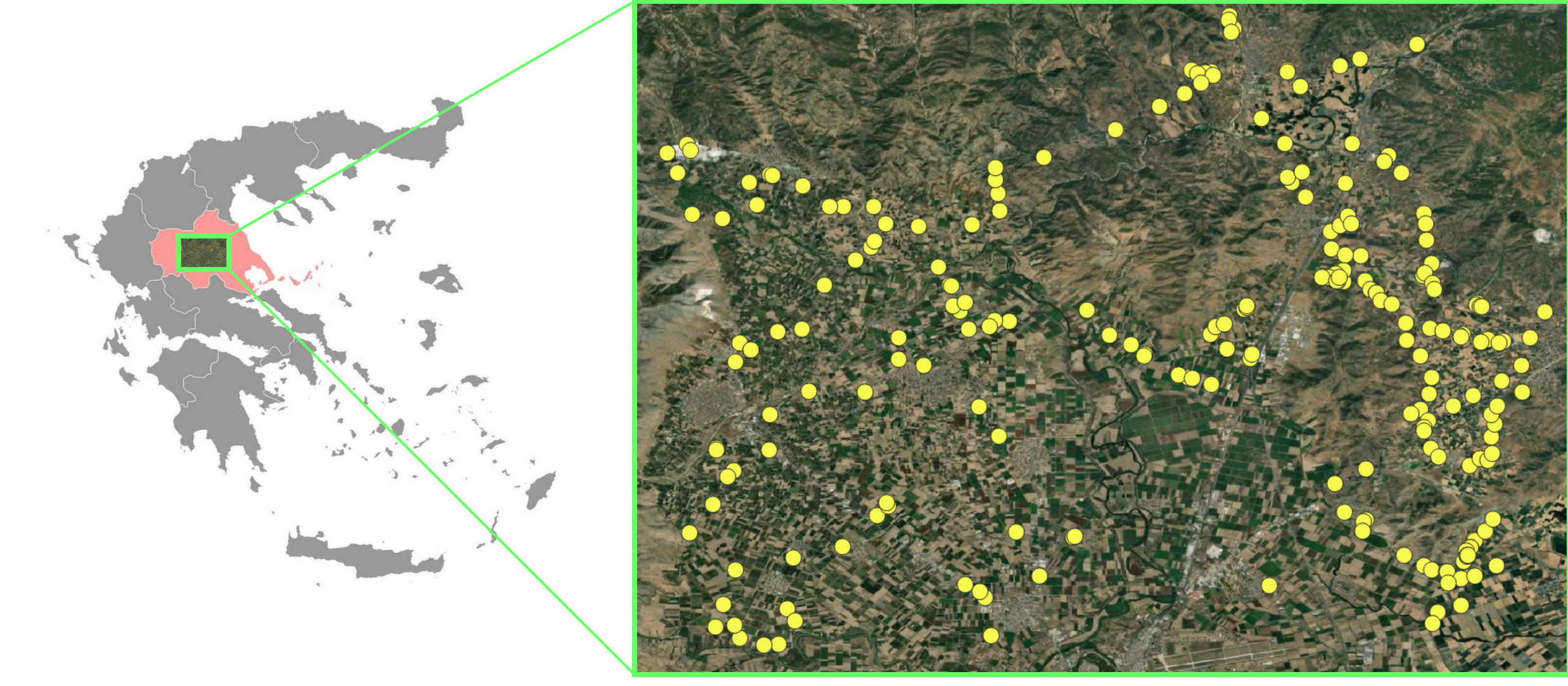}
    \caption{Our area of interest in Thessaly, Greece. Yellow points represent the fields where we identified a weed management method.}
    \label{fig:2.1}
\end{figure}

% \begin{comment} Our partner ELGO-DIMITRA \end{comment}
Data were collected from two agronomists by field visits in the region of Thessaly, Greece as depicted in Figure~\ref{fig:2.1} using ESRI's \textit{Survey123} mobile platform and stored in ArcGIS Online. 
% This software offers an online interface platform to store, view and share large volumes of data. 
The yellow points on the map represent the fields where we identified that one of the three aforementioned weed management methods was implemented in a period spanning 4 months (May-August) in 2024. 
In addition to the method, we also identified the orchard type of each inspected field.

Before proceeding to field identification and the extraction of satellite images, we conducted a preliminary exploratory data analysis to better understand the composition and characteristics of the dataset.  More specifically, as shown in Table ~\ref{tab:practices_per_crop}, there is a class imbalance, as \textit{Mowing} class occupies the majority of the weed management methods (61\%), while the rest of the classes - \textit{Tillage}, \textit{Chemical-spraying} and \textit{No practice}- represent 39\% of the whole dataset - 14\%, 13\% and 12\% respectively. In Table~\ref{tab:practices_per_crop}, it is clear that for all orchard types the majority of weed management practices implemented in the respective fields is \textit{Mowing}.

% \usepackage{booktabs}
% \begin{table}[]
% \centering
% \caption{Count of Weed Management methods per orchard}
% \label{tab:practices_per_crop}
% \begin{tabular}{@{}ccccc@{}}
% \toprule
% \textbf{Orchard}     & \textbf{Mowing} & \textbf{Tillage} & \textbf{Chemical-spraying} &\textbf{No practice} \\ \midrule
% Apricots &     29    &   1      &  1 &    0             \\ \midrule
% Peaches  &  29       &  2       &  6 &   2              \\ \midrule
% Almonds  &   42      &    25     & 18 & 7                 \\ \midrule
% Pears    &    11     &    0     &   1  &  1            \\ \midrule
% Olives   &   27      &      5   &   5  & 16              \\ \midrule
% Pistachios   &    3     &     0    &   0  &   1           \\ \midrule
% \textbf{Total}    &   \textbf{141}      &    \textbf{33}     &  \textbf{31}     &\textbf{27}            \\ \bottomrule
% \end{tabular}
% \end{table}

\subsection{Satellite Dataset} 
For the purpose of our analysis, we used two satellite systems, S2 and PS. S2, operated by the European Space Agency (ESA), provides multispectral imagery with a spatial resolution of up to 10 meters, and a temporal resolution of 5 days. PS, operated by PlanetLabs, offers daily multispectral imagery at a higher spatial resolution of 3 meters, allowing for more detailed observations at the field level. For the PS data analysis, we utilized the Surface Reflectance (SR) 8-band (8B) product, which is optimized to reduce the variability of spectral bands due to atmospheric effects. Table \ref{tab:spectral_bands} summarizes the bands extracted from the images of each satellite.

\begin{table}[ht!]
\centering
\caption{Count of Weed Management methods per orchard}
\label{tab:practices_per_crop}
\begin{tabular}{@{}ccccc@{}}
\toprule
\textbf{Orchard}     & \textbf{Mowing} & \textbf{Tillage} & \textbf{Chemical-spraying} &\textbf{No practice} \\ \midrule
Apricots &     29    &   1      &  1 &    0             \\ \midrule
Peaches  &  29       &  2       &  6 &   2              \\ \midrule
Almonds  &   42      &    25     & 18 & 7                 \\ \midrule
Pears    &    11     &    0     &   1  &  1            \\ \midrule
Olives   &   27      &      5   &   5  & 16              \\ \midrule
Pistachios   &    3     &     0    &   0  &   1           \\ \midrule
\textbf{Total}    &   \textbf{141}      &    \textbf{33}     &  \textbf{31}     &\textbf{27}            \\ \bottomrule
\end{tabular}
\end{table}

\begin{table}[ht!]
\centering
\caption{Spectral Bands of S2,and PS-8B}
\label{tab:spectral_bands}
\begin{tabular}{|c|l|c|}
\hline
\textbf{Satellite System} & \textbf{Band Description}\\ \hline
\multirow{8}{*}{S2} & Band 1 - Coastal Aerosol   \\ 
                            & Band 2 - Blue \\ 
                            & Band 3 - Green \\ 
                            & Band 4 - Red   \\
                            & Band 5 - Red Edge 1   \\
                            & Band 6 - Red Edge 2   \\
                            & Band 7 - Red Edge 3   \\
                            & Band 8 - Near-Infrared (NIR)   \\
                            & Band 8A - Narrow Near-Infrared (Narrow NIR)   \\
                            & Band 9 - Water Vapour   \\
                            & Band 10 - Shortwaved Infrared   \\
                            & Band 11 - Shortwaved Infrared 1 (SWIR1)   \\
                            & Band 12 - Shortwaved Infrared 2 (SWIR2)   \\
                            \hline

\multirow{8}{*}{PS-8B} & Band 1 - Coastal Blue  \\ 
                                    & Band 2 - Blue      \\ 
                                    & Band 3 - Green I     \\ 
                                    & Band 4 - Green    \\ 
                                    & Band 5 - Yellow    \\ 
                                    & Band 6 - Red   \\ 
                                    & Band 7 - Red Edge  \\ 
                                    & Band 8 - Near Infrared (NIR)  \\ \hline
\end{tabular}
\end{table}

\subsection{Feature Extraction}
 
In addition to the spectral bands obtained from the satellite collections, we calculated the Normalized Difference Vegetation Index (NDVI) as an extra feature for each satellite, as shown in Equation \ref{eq:NDVI_S2} (S2) and Equation \ref{eq:NDVI_PS} (PS 8-band product). NDVI was included because its normalized difference formula enhances sensitivity to vegetation changes (e.g., weed biomass reduction from mowing or soil disturbance from tillage) that raw spectral bands may miss, due to noise from soil brightness or shadows. To capture temporal dynamics, we also calculated for all features, the first-order differences (Equation \ref{eq:diff}) and rates of change (Equation ~\ref{eq:rate_of_change}), where $x_i$ represent the value of feature $x$ at the time $i$, and $t_{i+1} - t_i$, represent the days passed from last observation. 

\begin{comment}
NDVI’s normalized difference formula highlights vegetation-specific signals while suppressing noise from soil brightness or shadows—subtle effects that raw spectral bands may miss.
\end{comment}

\begin{comment}
NDVI was included alongside spectral bands to provide a vegetation-specific index known to improve separability in vegetation classification tasks.
\end{comment}

\begin{equation}
\label{eq:NDVI_S2}
NDVI_{S2} = \frac{B08 - B04}{B08 + B04}
\end{equation}

\begin{equation}
\label{eq:NDVI_PS}
NDVI_{PS8B} = \frac{B08 - B06}{B08 + B06}
\end{equation}

% \begin{comment}
    
\begin{equation}
\label{eq:diff}
\Delta x_i = x_{i+1} - x_i
\end{equation}

% \end{comment}

\begin{equation}
\label{eq:rate_of_change}
% \text{Rate of Change} 
ROC = \frac{x_{i+1} - x_i}{t_{i+1} - t_i}
\end{equation}

% \begin{table}[h!]
% \centering
% \caption{Spectral Bands of S2,and PS-8B}
% \label{tab:spectral_bands}
% \begin{tabular}{|c|l|c|}
% \hline
% \textbf{Satellite System} & \textbf{Band Description}\\ \hline
% \multirow{8}{*}{S2} & Band 1 - Coastal Aerosol   \\ 
%                             & Band 2 - Blue \\ 
%                             & Band 3 - Green \\ 
%                             & Band 4 - Red   \\
%                             & Band 5 - Red Edge 1   \\
%                             & Band 6 - Red Edge 2   \\
%                             & Band 7 - Red Edge 3   \\
%                             & Band 8 - Near-Infrared (NIR)   \\
%                             & Band 8A - Narrow Near-Infrared (Narrow NIR)   \\
%                             & Band 9 - Water Vapour   \\
%                             & Band 10 - Shortwaved Infrared   \\
%                             & Band 11 - Shortwaved Infrared 1 (SWIR1)   \\
%                             & Band 12 - Shortwaved Infrared 2 (SWIR2)   \\
%                             \hline

% \multirow{8}{*}{PS-8B} & Band 1 - Coastal Blue  \\ 
%                                     & Band 2 - Blue      \\ 
%                                     & Band 3 - Green I     \\ 
%                                     & Band 4 - Green    \\ 
%                                     & Band 5 - Yellow    \\ 
%                                     & Band 6 - Red   \\ 
%                                     & Band 7 - Red Edge  \\ 
%                                     & Band 8 - Near Infrared (NIR)  \\ \hline
% \end{tabular}
% \end{table}

\section{Methodology}

\begin{figure}[ht!]
    \centering
    \includegraphics[width=\columnwidth]{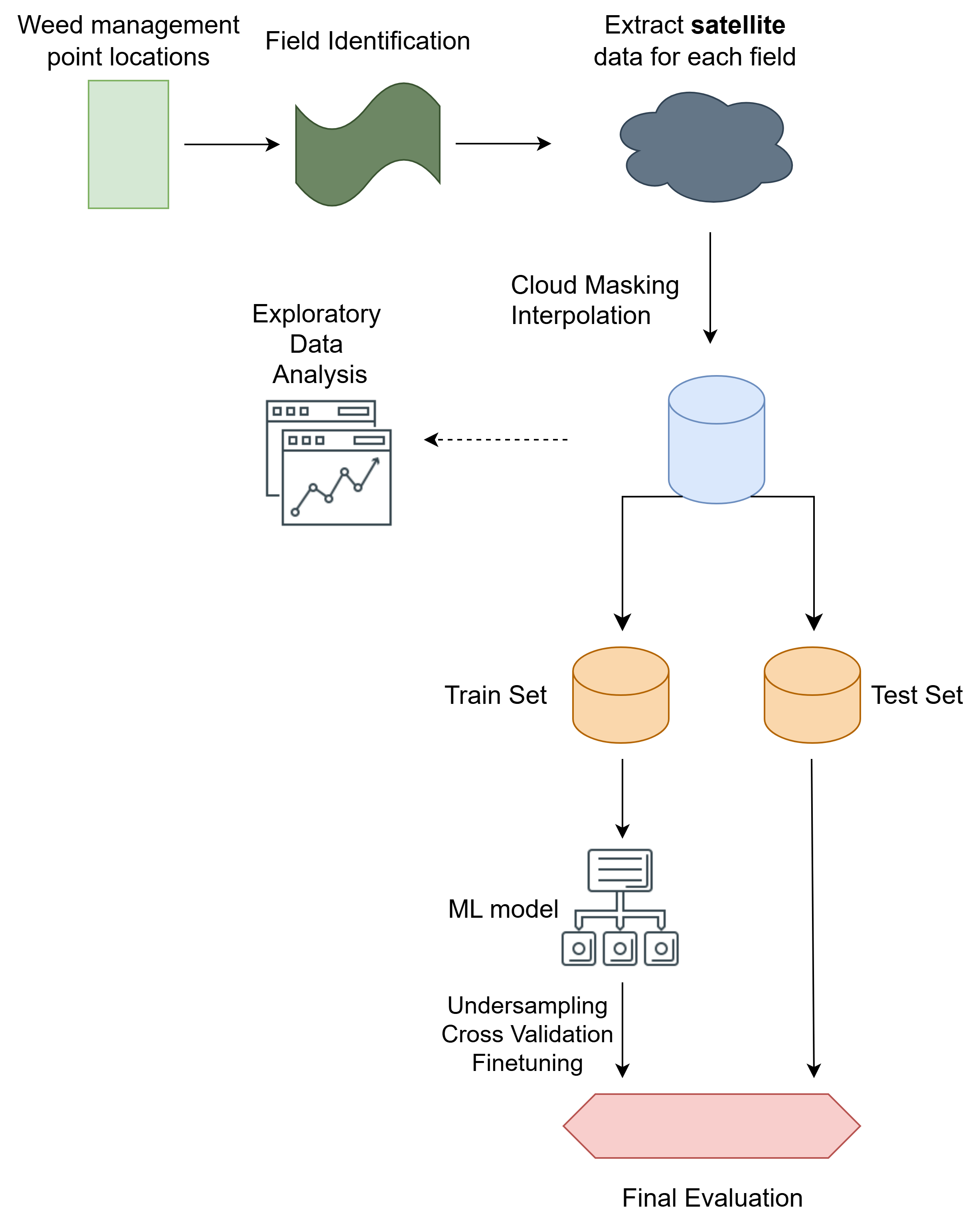}
    \caption{Overview of the methodology pipeline for weed management method classification. The process includes data collection, data pre-processing, exploratory data analysis, feature engineering, training a classification model and evaluating its performance.  }
    \label{fig:3.1}
\end{figure}

Our methodology pipeline consists of several steps that are summarized in Figure~\ref{fig:3.1}. The goal of our approach, is to predict which weed management method has been implemented in a field within the past 4 months, based on its historical satellite 4-months time series observations, which will serve as the input variable \textit{X}. Respectively, \textit{Y} represents the management method that took place in the field at least once during this 4-months period. In this study we do not focus on predicting the exact date that the method realized but rather on classifying the weed management method that was most likely applied. After collecting the weed management points locations, we visually delineated the agricultural field geometries, using available PS images. Utilizing the field geometries, we extracted satellite data for a 4-month period, from May to August, in the form of spectral bands time series. During these months, it is more common to implement a weed management method in Greece due to the active growth of weeds, facilitated by favorable climatic conditions such as higher temperatures and sufficient moisture.
% In order to tackle class imbalance, we randomly excluded $0.6\%$  of the majority class fields (i.e., fields that \textit{Mowing} took place) in the training dataset. 

In the next phase, we pre-processed the 2 different satellite datasets to ensure compatibility with ML algorithms. Particularly, we excluded images with greater than 0.5\% cloud coverage. In order to fill the gaps between the measurements and to ensure temporal compatibility for future observations, we applied 10-day linear interpolation to both datasets.

Following this process, we constructed 2 pixel-based datasets (S2 and PS) - each row represents a single pixel throughout 4 months - which will be then used as input for the classification algorithms. The approach and experimental setup employed to evaluate the effectiveness of these datasets for the problem of weed management methods classification is detailed in the following sections.

% \begin{figure}[]
%     \centering
%     \includegraphics[width=\columnwidth]{Figures/PNG/fig2.png}
%     \caption{Overview of the methodology pipeline for weed management method classification. The process includes data collection, data pre-processing, exploratory data analysis, feature engineering, training a classification model and evaluating its performance.  }
%     \label{fig:3.1}
% \end{figure}

\subsection{Classification approach}

The effectiveness of satellite data in classifying the aforementioned weed management methods is evaluated using three established ML algorithms: Random Forest (RF), Extreme Gradient Boosting (XGB), and K-Nearest Neighbors (KNN).

Our approach employs parcel-based classification, which aggregates pixel features to the field level based on a shared geometric boundary. Specifically, the mean, median, and standard deviation of all features within each field are calculated, with each field representing a single instance. This aggregation simplifies the data and reduces noise from pixel-level variations, such as those caused by tree canopies. Overall, the PS dataset, with its higher spatial resolution (3m), is expected to yield improved results compared to the lower resolution (10m) S2, as it provides a greater number of pixel instances for analysis.

\subsection{Experimental Setup}

% To evaluate the performance of the aforementioned parcel-based classification approach for our task, we conducted the proposed methodology for 2 different experiments, utilizing the 2 datasets created. For all the cases, the modeling phase is the same: We left out 20\% of the total fields for testing and we utilized the rest for the model training. The experiments are structured as follows:
% For each constructed satellite dataset, we train three standard machine learning classification models, Random Forest, XGB and KNN, and evaluate them separately for each dataset. 
Based on this parcel-based approach, we conducted two independent experiments, one for each created satellite collection. A stratified random split was applied, reserving the same 20\% of fields testing,  while the remaining 80\% were used for training the ML algorithms. In order to tackle class imbalance, we randomly excluded $0.6\%$  of the majority class fields (i.e., fields that \textit{Mowing} took place) in the training dataset. Then, using the training set we performed cross-validation and finetuned the models' parameters.
We lastly evaluated the performance of the finetuned models on the test set, using precision, recall, F1 score for each class and average weighted F1 score as metrics. 
% More specifically, the mean of each performance metric was estimated through a 5-fold cross validation. 
% For the evaluation of the models, the support samples represented by 11 fields that belong to \textit{Mowing} (Class 0), 7 fields belong to \textit{Tillage}  (Class 1), 6 belong to \textit{Chemical-spraying} (Class 2) and 5 belong to \textit{No practice} (Class 3).
The support samples included 29 fields for \textit{Mowing}, 7 for \textit{Tillage}, 7 for \textit{Chemical-spraying}, and 6 for \textit{No practice}.

% The purpose of these experiments is to provide insights of the strengths and limitations of each satellite data source for the task of weed management method classification.

\section{Results and Discussion}

Table ~\ref{tab:model_performance} provides performance metrics of the finetuned models for the parcel-based classification, evaluated on the reserved test set, across the four classes: \textit{Mowing} (MO), \textit{Tillage} (TL), \textit{Chemical-spraying} (CS) and \textit{No practice} (NP).

\begin{table}[ht]
\centering
\caption{Classification evaluation on test set for RF, XGB, and KNN models on S2 and PS datasets.}
\footnotesize
\setlength{\tabcolsep}{4pt} % Adjust column padding
\renewcommand{\arraystretch}{1.2} % Adjust row height
\begin{tabular}{ccccccc}
\toprule
\textbf{Data} & \textbf{Model} & \textbf{Class} & \textbf{Precision} & \textbf{Recall} & \textbf{F1 Score} & \textbf{Weighted F1}\\
\midrule

\multirow{9}{*}{S2} 
    & \multirow{4}{*}{RF}  & MO & 0.63 & \textbf{0.83} & 0.72 & \multirow{4}{*}{0.49} \\
    &                      & TL & 0.2 & 0.14 & 0.17 & \\
    &                      & CS & 0.0 & 0.0 & 0.0 & \\
    &                      & NP & 0.4 & 0.33 & 0.36 & \\

\cline{2-7}
    & \multirow{4}{*}{XGB} & MO & 0.60 & 0.72 & 0.66 & \multirow{4}{*}{0.45} \\
    &                      & TL & 0.14 & 0.14 & 0.14 & \\
    &                      & CS & 0.0 & 0.0 & 0.0 & \\
    &                      & NP & 0.33 & 0.33 & 0.33 & \\

\cline{2-7}
    & \multirow{4}{*}{KNN} & MO & 0.64 & 0.79 & 0.71 & \multirow{4}{*}{0.48} \\
    &                      & TL & 0.17 & 0.29 & 0.21 & \\
    &                      & CS & \textbf{1.0} & 0.14 & 0.25 & \\
    &                      & NP & 0.0 & 0.0 & 0.0 & \\

\midrule
\midrule

\multirow{9}{*}{PS} 
    & \multirow{4}{*}{RF}  & MO & \textbf{0.69} & 0.76 & \textbf{0.72} & \multirow{4}{*}{\textbf{0.57}} \\
    &                      & TL & \textbf{0.44} & \textbf{0.57}  & \textbf{0.5} & \\
    &                      & CS & 0.33 & \textbf{0.29} & \textbf{0.31} & \\
    &                      & NP & \textbf{0.5} & 0.17 & 0.25 & \\

\cline{2-7}
    & \multirow{4}{*}{XGB} & MO & 0.62 & 0.62 & 0.62 & \multirow{4}{*}{0.49} \\
    &                      & TL & 0.33 & 0.29 & 0.31 & \\
    &                      & CS & 0.25 & 0.29 & 0.27 & \\
    &                      & NP & 0.33 & 0.33 & 0.33 & \\

\cline{2-7}
    & \multirow{4}{*}{KNN} & MO & 0.65 & 0.69 & 0.67 & \multirow{4}{*}{0.54} \\
    &                      & TL & 0.43 & 0.43 & 0.43 & \\
    &                      & CS & 0.33 & 0.29 & 0.31 & \\
    &                      & NP & 0.4 & \textbf{0.33} & \textbf{0.36} & \\

\bottomrule
\end{tabular}
\label{tab:model_performance}
\end{table}

% Sentinel-2 data demonstrated poor performance in comparison with the PlanetScope data, likely due to to its lower spatial resolution. \textit{Mowing} achieved the highest F1 score on all models, suggesting that mowing practice implementation impacts the field more distinctly, making it easier for the models to identify. However, \textit{Tillage} and \textit{Chemical-spraying}, were more challenging to classify, possibly because the effects of these practices are more subtle and harder to detect through Sentinel-2 resolution. 

Models trained on S2 data demonstrated generally lower performance compared to PS data, likely due to its lower spatial resolution. \textit{Mowing} achieved the highest F1 score on all S2-trained models, suggesting that mowing practice implementation impacts the field more distinctly, making it easier for the models to identify, as RF model suggests. However, the rest of the practices, were more challenging to classify, possible because the effects of these practices are more subtle and harder to detect through S2 resolution.

% PlanetScope data performed noticeably better that Sentinel-2, as the higher spatial resolution (3m) allowed for a more detailed discrimination of the field-level features. All classes achieved higher F1 scores. 
% \textit{Tillage} displayed the best results,while \textit{Mowing} and \textit{No practice} showed noticeable improvement. However, \textit{Chemical-spraying}, despite the boost on metrics performance, remained the most challenging to classify. This can be attributed to the subtle impact of this practice to the field in relation to the use of satellite imagery to detect it. 

PS data models performed noticeably better than S2, as the higher spatial resolution (3m) allowed for a more detailed discrimination of the field-level features. RF outperformed the other models in terms of weighted F1 score, while KNN noted competitive results, achieving the highest F1 score for \textit{No practice}. \textit{Chemical-spraying}, despite the boost in metrics performance, remained the most challenging to classify. This can be attributed to the subtle impact of this practice to the field in relation to the use of satellite imagery to detect it.

% All classes achieved higher F1 scores. 
% \textit{Tillage} displayed the best results,while \textit{Mowing} and \textit{No practice} showed noticeable improvement. However, \textit{Chemical-spraying}, despite the boost on metrics performance, remained the most challenging to classify. This can be attributed to the subtle impact of this practice to the field in relation to the use of satellite imagery to detect it. 

\begin{figure}[ht!]
    \centering
    \includegraphics[width=\columnwidth]{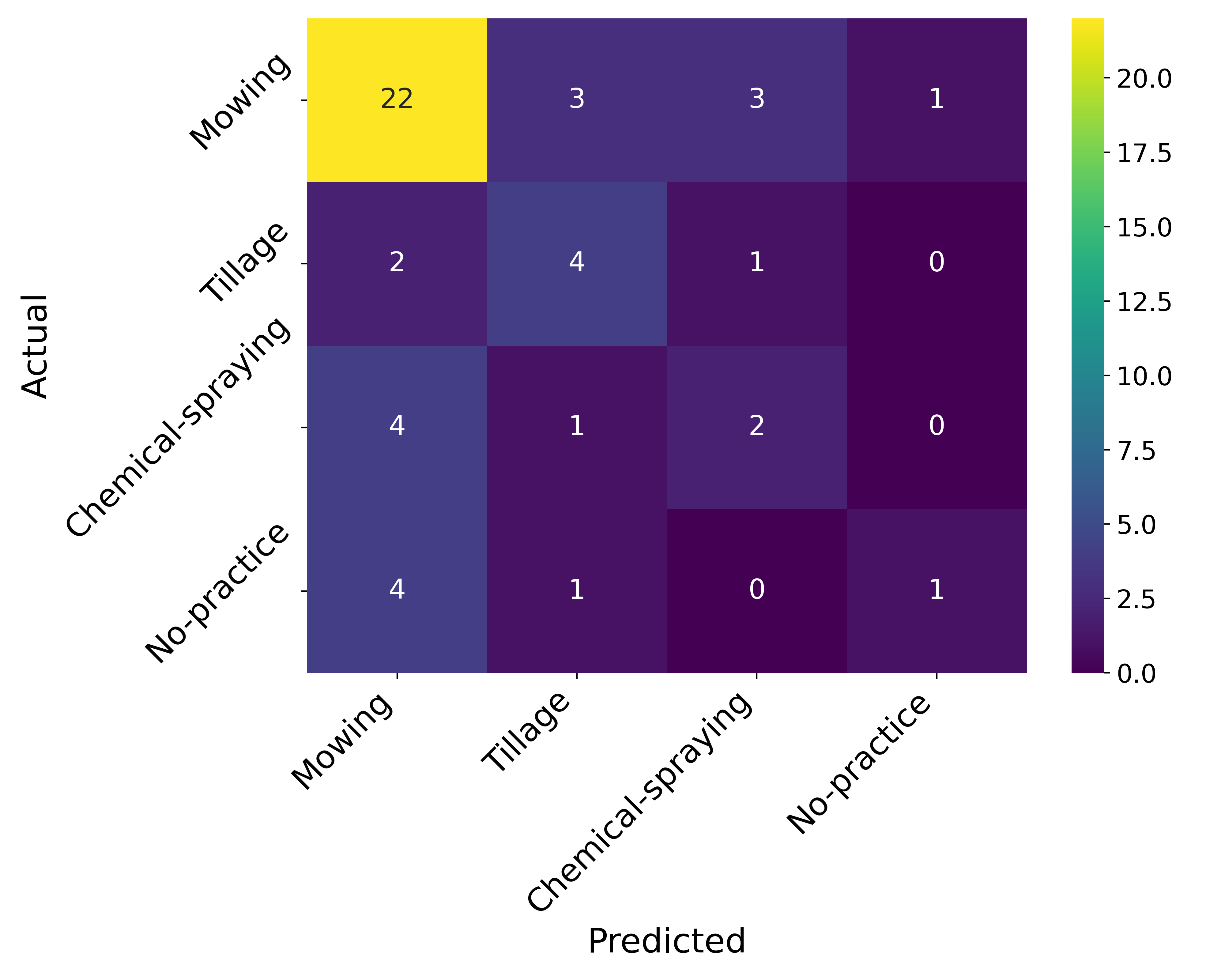}
    \caption{Confusion matrix for the best RF model trained and evaluated on PS data. Warmer colors indicate higher agreement (i.e., more correct predictions), while colder colors represent lower agreement.}
    \label{fig:cfm}
\end{figure}

The confusion matrix of the best performing model on Figure~\ref{fig:cfm} highlights that the model most often correctly predicts \textit{Mowing} and \textit{Tillage}, as their spectral impact on the field is more discrete. \textit{Chemical-spraying} and \textit{No practice} class were poorly predicted by the same model, as half or more of the support fields for these classes, were classified as 
\textit{Mowing}, most likely due their similarities on spectral patterns and lack of training data.
As weighted F1 score - $0.57$ - suggests, the best performing model is RF combined with the PS data.

% \textit{Mowing} and \textit{Tillage} according to relatively high F1 scores, $0.72$ and $0.5$ respectively, are simpler to classify since mowing causes reduction in vegetation height and tillage result in the exposure of bare soil.However, the performance of the models is limited by the small amount of data (232 fields). To address these challenges, enhancing the dataset with a more diverse number of fields during the 2025 campaign, alongside with a combination of other sources like Sentinel-1, could significantly contribute to this weed management mapping approach. In general, RF showed superior performance in classifying \textit{Mowing} and \textit{Chemical-spraying}, while XGB outperformed the other models, when dealing with \textit{Tillage} and \textit{No practice}.

\textit{Mowing} and \textit{Tillage} according to relatively high F1 scores, $0.72$ and $0.5$ achieved by the RF - trained on PS data - respectively, are simpler to classify since mowing causes reduction in vegetation height and tillage result in the exposure of bare soil. 
Compared to studies like ~\cite{rouault2023grass}
, which achieved $57\%$ accuracy for simpler classifications of grassed vs non-grassed plots using Pleiades images (30cm spatial resolution), and ~\cite{kandwal2009discriminating}
, with $69\%$ accuracy using NDVI for invasive species detection, our results (up to $0.57$ weighted F1 score with PS) are promising, taking into account the complexity of the problem.
However, the performance of the models is limited by the small amount of data (232 fields), and the uneven representation of weed management practices.
To address these challenges, enhancing the dataset with a more diverse number of fields during the 2025 campaign, alongside with a combination of other sources like Sentinel-1, could significantly contribute to this weed management mapping approach.

\section{Conclusion \& Future Work}

% Future research should focus on data enrichment, in order to include more weed management methods, feature engineering techniques, such as pixel-filtering, utilizing more advanced machine learning models and testing data fusion strategies, to further enhance classification accuracy. For future work,more advanced techniques like tree canopies pixel filtering might help the machine learning models to better distinguish between classes.

The study explores the utilization of satellite imagery and ML to map weed management methods in orchards. Our findings, demonstrate the potential of high resolution PS imagery for this task, outperforming S2. While challenges remain, such as class imbalance and subtle impact of some methods though satellite imagery,  the results highlight the potential effectiveness of EO data in weed management applications on larger scale. Future research should focus on data enrichment, in order to include additional weed management methods, applying innovative feature engineering techniques, such as pixel-filtering for tree canopies, utilizing more advanced ML models and testing data fusion strategies, to further enhance classification accuracy. In addition, satellite data fusion could be employed, in order to combine the advantages of both platforms.

\section{Acknowledgements}
This work was supported by the project "Climaca" (ID: 16196) which is carried out within the framework of the National Recovery and Resilience Plan Greece 2.0, funded by the European Union – NextGenerationEU. This study includes the use of PlanetScope imagery provided through the agreement between Wageningen University \& Research (WUR) and Planet Labs PBC. The data were used under the “Education and Research Rights” license for noncommercial scientific research. Planet Labs PBC is gratefully acknowledged as the source of the satellite imagery.

\small
\bibliographystyle{IEEEtranN}
\bibliography{ref}

% Generated by IEEEtranN.bst, version: 1.14 (2015/08/26)
\begin{thebibliography}{14}
\providecommand{\natexlab}[1]{#1}
\providecommand{\url}[1]{#1}
\csname url@samestyle\endcsname
\providecommand{\newblock}{\relax}
\providecommand{\bibinfo}[2]{#2}
\providecommand{\BIBentrySTDinterwordspacing}{\spaceskip=0pt\relax}
\providecommand{\BIBentryALTinterwordstretchfactor}{4}
\providecommand{\BIBentryALTinterwordspacing}{\spaceskip=\fontdimen2\font plus
\BIBentryALTinterwordstretchfactor\fontdimen3\font minus \fontdimen4\font\relax}
\providecommand{\BIBforeignlanguage}[2]{{%
\expandafter\ifx\csname l@#1\endcsname\relax
\typeout{** WARNING: IEEEtranN.bst: No hyphenation pattern has been}%
\typeout{** loaded for the language `#1'. Using the pattern for}%
\typeout{** the default language instead.}%
\else
\language=\csname l@#1\endcsname
\fi
#2}}
\providecommand{\BIBdecl}{\relax}
\BIBdecl

\bibitem[Monteiro and Santos(2022)]{monteiro2022sustainable}
A.~Monteiro and S.~Santos, ``Sustainable approach to weed management: The role of precision weed management,'' \emph{Agronomy}, 2022.

\bibitem[Bajwa et~al.(2015)Bajwa, Mahajan, and Chauhan]{bajwa2015nonconventional}
A.~A. Bajwa, G.~Mahajan, and B.~S. Chauhan, ``Nonconventional weed management strategies for modern agriculture,'' \emph{Weed science}, vol.~63, no.~4, pp. 723--747, 2015.

\bibitem[Parker(2003)]{parker2003water}
R.~Parker, ``Water conservation, weed control, go hand in hand,'' 2003.

\bibitem[Imanni et~al.(2023)Imanni, Harti, Bachaoui, Mouncif, Eddassouqui, Hasnai, and Zinelabidine]{Imanni2023Multispectral}
H.~S.~E. Imanni, A.~E. Harti, E.~M. Bachaoui, H.~Mouncif, F.~Eddassouqui, M.~A. Hasnai, and M.~I. Zinelabidine, ``Multispectral {UAV} data for detection of weeds in a citrus farm using machine learning and {Google} {Earth} {Engine}: Case study of {Morocco},'' \emph{Remote Sensing Applications Society and Environment}, vol.~30, pp. 100\,941--100\,941, feb 22 2023.

\bibitem[Esposito et~al.(2021)Esposito, Crimaldi, Cirillo, Sarghini, and Maggio]{Esposito2021Drone}
M.~Esposito, M.~Crimaldi, V.~Cirillo, F.~Sarghini, and A.~Maggio, ``Drone and sensor technology for sustainable weed management: a review,'' \emph{Chemical and Biological Technologies in Agriculture}, vol.~8, no.~1, mar 25 2021.

\bibitem[Sishodia et~al.(2020)Sishodia, Ray, and Singh]{sishodia2020applications}
R.~P. Sishodia, R.~L. Ray, and S.~K. Singh, ``Applications of remote sensing in precision agriculture: A review,'' \emph{Remote sensing}, vol.~12, no.~19, p. 3136, 2020.

\bibitem[PBC(2024--)]{planet2024}
\BIBentryALTinterwordspacing
P.~L. PBC, ``Planet application program interface: In space for life on earth,'' Planet, 2024--. [Online]. Available: \url{https://api.planet.com}
\BIBentrySTDinterwordspacing

\bibitem[Mudereri et~al.(2019)Mudereri, Dube, Adel-Rahman, Niassy, Kimathi, Khan, and Landmann]{mudereri2019comparative}
B.~Mudereri, T.~Dube, E.~Adel-Rahman, S.~Niassy, E.~Kimathi, Z.~Khan, and T.~Landmann, ``A comparative analysis of planetscope and sentinel sentinel-2 space-borne sensors in mapping striga weed using guided regularised random forest classification ensemble,'' \emph{The International Archives of the Photogrammetry, Remote Sensing and Spatial Information Sciences}, vol.~42, pp. 701--708, 2019.

\bibitem[Tsardanidis et~al.(2025)Tsardanidis, Koukos, Sitokonstantinou, Drivas, and Kontoes]{tsardanidis_grasslands}
I.~Tsardanidis, A.~Koukos, V.~Sitokonstantinou, T.~Drivas, and C.~Kontoes, ``Cloud gap-filling with deep learning for improved grassland monitoring,'' \emph{Computers and Electronics in Agriculture}, vol. 230, p. 109732, 2025.

\bibitem[Gao et~al.(2018)Gao, Liao, Nuyttens, Lootens, Vangeyte, Pi{\v{z}}urica, He, and Pieters]{gao2018fusion}
J.~Gao, W.~Liao, D.~Nuyttens, P.~Lootens, J.~Vangeyte, A.~Pi{\v{z}}urica, Y.~He, and J.~G. Pieters, ``Fusion of pixel and object-based features for weed mapping using unmanned aerial vehicle imagery,'' \emph{International journal of applied earth observation and geoinformation}, vol.~67, pp. 43--53, 2018.

\bibitem[Su et~al.(2022)Su, Yi, Coombes, Liu, Zhai, McDonald-Maier, and Chen]{su2022spectral}
J.~Su, D.~Yi, M.~Coombes, C.~Liu, X.~Zhai, K.~McDonald-Maier, and W.-H. Chen, ``Spectral analysis and mapping of blackgrass weed by leveraging machine learning and uav multispectral imagery,'' \emph{Computers and Electronics in Agriculture}, vol. 192, p. 106621, 2022.

\bibitem[Fragassa et~al.(2023)Fragassa, Vitali, Emmi, and Arru]{fragassa2023new}
C.~Fragassa, G.~Vitali, L.~Emmi, and M.~Arru, ``A new procedure for combining uav-based imagery and machine learning in precision agriculture,'' \emph{Sustainability}, vol.~15, no.~2, p. 998, 2023.

\bibitem[Rouault et~al.(2023)Rouault, Courault, Pouget, Flamain, Lopez-Lozano, Doussan, Debolini, and McCabe]{rouault2023grass}
P.~Rouault, D.~Courault, G.~Pouget, F.~Flamain, R.~Lopez-Lozano, C.~Doussan, M.~Debolini, and M.~McCabe, ``Grass cover, tree density, and leaf development of mediterranean orchards from high resolution data,'' \emph{The International Archives of the Photogrammetry, Remote Sensing and Spatial Information Sciences}, vol.~48, pp. 1531--1536, 2023.

\bibitem[Kandwal et~al.(2009)Kandwal, Jeganathan, Tolpekin, and Kushwaha]{kandwal2009discriminating}
R.~Kandwal, C.~Jeganathan, V.~Tolpekin, and S.~Kushwaha, ``Discriminating the invasive species,‘lantana’using vegetation indices,'' \emph{Journal of the Indian Society of Remote Sensing}, vol.~37, pp. 275--290, 2009.

\end{thebibliography}

\end{document}